\documentclass{article} 
\usepackage{iclr2021_conference,times}

\usepackage{amsmath,amsfonts,bm}

\def\eqref#1{equation~\ref{#1}}

\def\1{\bm{1}}

\DeclareMathAlphabet{\mathsfit}{\encodingdefault}{\sfdefault}{m}{sl}
\SetMathAlphabet{\mathsfit}{bold}{\encodingdefault}{\sfdefault}{bx}{n}

\usepackage{hyperref}
\usepackage{url}
\usepackage{graphicx}

\title{Street to Cloud: Improving Flood Maps With Crowdsourcing and Semantic Segmentation}

\iclrfinalcopy
\author{Veda Sunkara \\
Cloud to Street\\
New York, NY 112131, USA\\
\texttt{veda@cloudtostreet.info} \\
\And
Matthew Purri\\
Rutgers University\\
New Brunswick, NJ 08854, USA\\
\texttt{matthew.purri@rutgers.edu} \\
\And
Bertrand Le Saux \& Jennifer Adams \\
European Space Agency, ESRIN, I-00044 \\
Frascati (Rome), Italy \\
\texttt{\{Bertrand.Le.Saux,Jennifer.Adams\}@esa.int} \\
}

\begin{document}

\maketitle

\begin{abstract}
To address the mounting destruction caused by floods in climate-vulnerable regions, we propose Street to Cloud, a machine learning pipeline for incorporating crowdsourced ground truth data into the segmentation of satellite imagery of floods. We propose this approach as a solution to the labor-intensive task of generating high-quality, hand-labeled training data, and demonstrate successes and failures of different plausible crowdsourcing approaches in our model. Street to Cloud leverages community reporting and machine learning to generate novel, near-real time insights into the extent of floods to be used for emergency response. 
\end{abstract}
\section{Introduction}

The frequency and magnitude of flooding are increasing at an alarming rate~\citep{undrr_2015}, affecting growing populations of climate-vulnerable people. Flooding affects more people than any other environmental hazard and hinders sustainable development~\citep{ hallegatte_vogt-schilb_bangalore_rozenberg_2017, 2018_review_of_disaster_events_2019}, and research consistently shows that relative property loss for floods are highest in places of social vulnerability~\citep{tellman_schank_schwarz_howe_de_sherbinin_2020}.

Nowcasting flood extents enables decision makers, relief agencies, and citizens to make informed decisions and provide direct relief where it is needed most. Optical, radar, and microwave satellites make it possible to remotely create scalable, low-cost, and high-quality flood maps and impact assessments. However, there are significant challenges to flood mapping, monitoring, and analyzing based on satellite data. Unique challenges arise from infrequent revisit times, varying resolutions across satellites, adverse and obscuring weather conditions, and difficult to parse images of urban areas where most of the world's population and assets are concentrated. 

Most existing algorithms to process these images, machine learning or otherwise, use finely annotated data that often requires remote sensing expertise to generate. Traditional, threshold-based remote sensing often requires a nontrivial amount of manual quality assurance and parameter tuning from domain experts.

\begin{figure}[h!]
    \begin{center}
    \includegraphics[width=\textwidth]{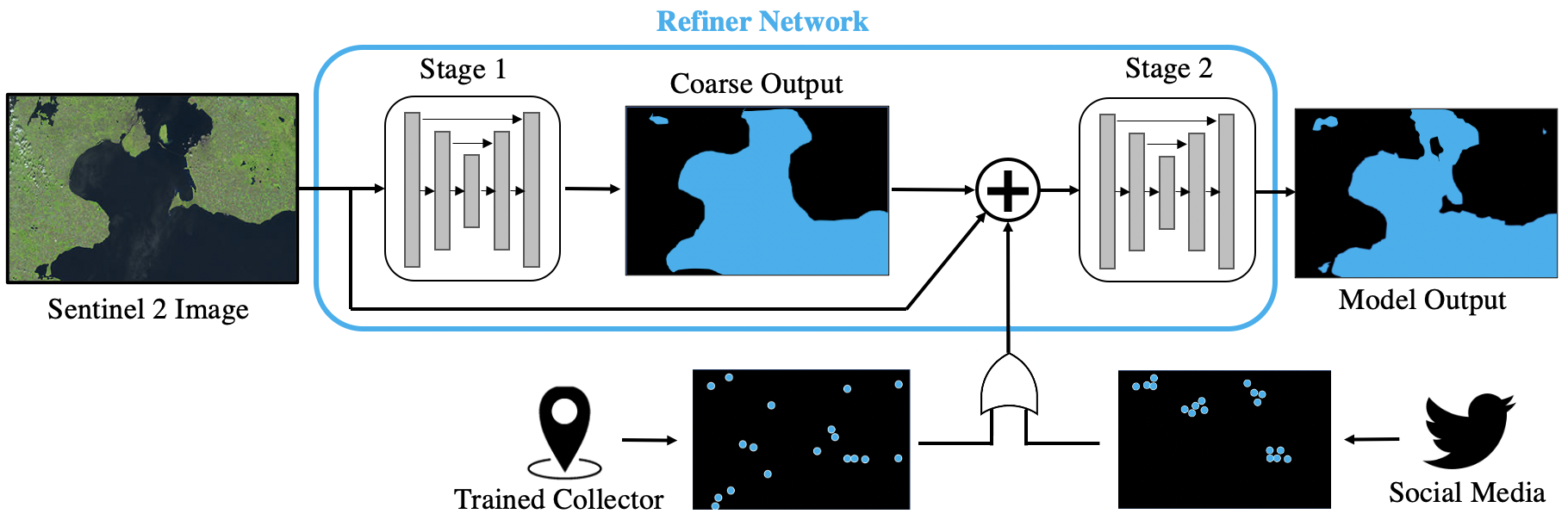}
    \caption{The inference pipeline of our model. The two-stage model first generates a segmentation mask from Sentinel-2 imagery in Stage 1, and then combines Sentinel-2 imagery, initial coarse output, and crowdsourced points in Stage 2 to generate the final segmentation mask. Points collected from either a Trained Collector or Social Media model can be used interchangeably in this model.}
    \label{fig:pipeline}
    \end{center}
\end{figure}

In an effort to develop an algorithm that not only addresses these data issues but also directly engages the communities affected in disaster reporting, we propose a methodology for using crowd-sourced data and simplified flood masks to train a semantic segmentation model to generate high quality flood masks. Using Cloud to Street's Sen1Floods11 dataset~\citep{Bonafilia_2020_CVPR_Workshops} of high-quality hand-labeled Sentinel-2 imagery, we created a dataset of simplified flood masks and synthetic crowdsourced data points. These masks are intended to be simple to generate even without remote sensing expertise, and therefore can be generated easily and at scale. Our synthetic crowdsourced data mirrors two plausible scenarios for aggregating data from the community: passive social media scraping and active data collection by community members or trained data collectors. Leveraging dense and sparse data at the same time is a challenge for segmentation networks that we tackle by adopting a two-stage process (see Figure~\ref{fig:pipeline}) in which the second stage is inspired by continual learning. After training our network using these two approaches, we benchmark our results against the models trained on purely hand-labeled and purely simplified training masks.

We expect this research to allow us to provide high quality, rapidly available flood maps for evacuation and aid. In the case of some urban areas, crowdsourcing will enable us to verify flooding on a street-by-street level where remote sensing data alone cannot. Flood waters recede quickly, sometimes before there is a satellite overpass or the clouds clear, rendering optical remote sensing data insufficient for flood detection. Similarly, radar data, which can map water through clouds, is often very noisy in urban areas as signals can bounce off buildings. With street-level crowdsourcing and machine learning, we can train models to do necessary initial inundation detection and compensate for challenges when only using satellite data. 

In this work we provide a dataset of simplified water masks of flood events, built off of Sen1Floods11, as well as a dataset of synthetic crowdsourced data for each event in a number of plausible collection scenarios. We present Street to Cloud, a multi-modal model framework which combines satellite imagery and in-situ, crowdsourced data in a segmentation and refiner network to produce nowcast flood extent maps for monitoring, aid, and disaster relief.

\section{Related Work}

Prior research using the Sen1Floods11 dataset has demonstrated gains in using Fully Convolutional Neural Networks (FCNNs) to segment Sentinel-2 imagery of floods over threshold-based  methods~\citep{Bonafilia_2020_CVPR_Workshops}. Of all the training strategies discussed, the most successful approach required training the network on hand-labeled images of flood events which we use in our work. Other approaches, such as DeepWaterMap~\citep{8913594}, generate water segmentation maps of Landsat-7 imagery with global surface water labels.

Multi-modal approaches to semantic segmentation of remotely sensed imagery build off of significant prior work geolocalizing data and incorporating crowdsourcing into disaster tracking. Efforts to geolocalize street-view imagery have shown promising results using feature matching between aerial and ground data~\citep{regmi2019bridging}. The methods described can be used to identify a photographer's angle and location when parsing crowdsourced images. Other work has delved further into flood classification from social media imagery as well as separately in satellite imagery~\citep{ceur}, providing promising baselines for inferring ground truth data from social media images. There are examples of incorporating crowdsourcing into flood monitoring, including to assess flood depth~\citep{hultquist_cervone_2020} and for interactive flood modeling~\citep{gebremedhin_basco_carrera_jonoski_iliffe_winsemius_2020}. 

Exploration into iterative segmentation using human-in-the-loop annotation~\citep{xu2016deep, Lenczner_2020} suggests potential gains to be made using ground-truth verified data in addition to initial segmentation masks.

\section{Methods}

We generated two new datasets to train our model: coarse water masks of flood events and corresponding synthetic crowdsourced data points. To generate the coarse water masks, we used hand labeled Sentinel-2 imagery from the Sen1Floods11 dataset and simplified the water masks using a Gaussian blur with a large kernel. To generate the synthetic crowdsourced data, we sought to emulate two plausible approaches to data collection. The first is to emulate social media scraping, in which we anticipate a significant number of data points coming from the same region in an event (e.g. a populated street, community center). These points have low dispersion. The second is to emulate more spread out crowdsourced data that could be obtained by contractors walking the perimeter of an event and providing data points at regular intervals. These points have high dispersion. 

The total number of points per image is between 20 and 50, which makes up roughly 0.02\% of the total points in the image. We sample these points from the edge of the water annotation in Sen1Floods11, either in clusters or with a higher dispersion factor to emulate these two scenarios. In addition, we incorporate varied amounts of noise into the data to emulate geolocalization and user errors (e.g. distance from a reported flood event boundary). The range of simulated noise from a GPS sensor is 0 to 50 and 0 to 100 meters for low and high noise scenarios, respectively. The points are aligned with the other data modalities by projecting the generated points onto a blank image.

We introduce a two-stage segmentation network to utilize both multispectral Sentinel-2 imagery and crowdsourced points which we call the Refiner Network. The first stage of the network is tasked with generating a course water mask as shown in Figure~\ref{fig:pipeline}. The second stage refines on the coarse prediction by receiving crowdsourced points, the coarse mask, and multispectral imagery to generate a fine-grained output. We compare our Refiner Network to a standard UNet model~\citep{ronneberger2015UNet}.

\section{Results}
We assess model performance with pixel accuracy and mean intersection over union (mIoU). We benchmark against training our segmentation network on coarse labels, on finely annotated labels, and then on coarse labels with varying permutations of synthetic crowdsourced data points and noise. 

Our two-stage Refiner segmentation network outperforms the standard UNet architecture for both metrics on coarse and fine annotation levels as shown in Table~\ref{tab:main_results}. The difference between these models is greater when trained on the coarse data than when trained on the fine data, suggesting that the refiner model is able to take advantage of more precise annotations. The refiner model, when trained with a small number of synthetic points added to the coarse annotations, nears the performance of the UNet model trained on fine annotations. 
\begin{table}[h!]
    \centering
    \begin{tabular}{@{}|l|l|l|l|@{}}
        \hline
        Model     & Training Labels    & Acc & mIoU \\ \hline
        UNet      & Coarse        &  95.2   &  53.8    \\
        Refiner   & Coarse        &  95.6   &  56.5    \\ \hline
        Refiner   & Coarse+Points &  97.2   &  61.8    \\ \hline
        \textit{UNet}      & \textit{Fine}          &  \textit{97.0}   &  \textit{62.4}    \\
        \textit{Refiner}   & \textit{Fine }         &  \textit{98.1}   &  \textit{64.9 }   \\
        \hline
    \end{tabular}
    \caption{Comparison of model performance across training data annotation granularity and inclusion of crowdsourced points. Coarse training labels are created using Gaussian blur of hand labeled data, fine training labels are reference hand labeled data, and crowdsourced points are synthetically generated. We represent the best performing crowdsourcing scenario, as discussed further in Table~\ref{tab:points_ablation}.}
    \label{tab:main_results}
\end{table}

\begin{figure}[h!]
    \centering
        \includegraphics[width=\textwidth, height=5cm]{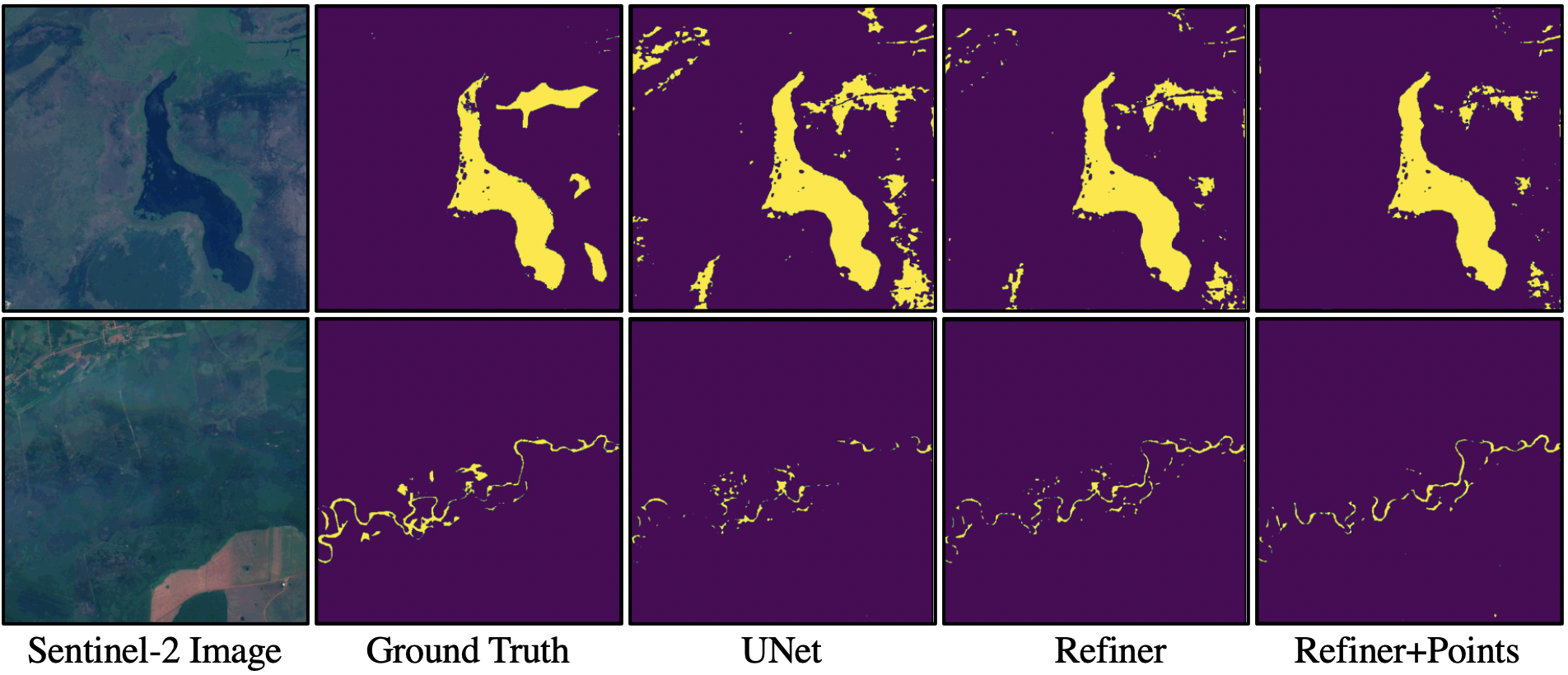}
        \caption{Qualitative results of the UNet, Refiner, and Refiner+Points models trained on coarse annotations. The Refiner+Points model appears to reduce false positives that other models generate.}
    \label{fig:qual_result}
\end{figure}

In Figure~\ref{fig:qual_result}, we show the qualitative improvement of utilizing crowdsourced points. The addition of crowdsourced points during training improves the model's ability to localize small water areas such as streams or flooded pools. In the bottom of Figure~\ref{fig:qual_result}, notice the Refiner model with points generated the most complete river compared to the other models. The low cost and minimal additional data from crowdsourced points greatly improves the performance of the network, and nears the upper bound performance of the UNet model trained on more expensive and time consuming annotations.

We then analyze what form of crowdsourced points improve segmentation performance. In Table~\ref{tab:points_ablation}, we compare crowdsourced points generated from a 'social media' (SM)  and 'trained data collector' (TDC) model, or low and high dispersion points respectively, along the fine annotation border. In Table~\ref{tab:points_ablation}, highly dispersed points result in higher model performance compared to the less dispersed points. In any situation the addition of crowdsourced points improves the performance of the refiner model over the baseline trained purely on coarse labels. Highly dispersed points with minimal noise produce the greatest improvement over the coarse, no point baseline. The importance of sensor noise affects each model differently. More severe sensor noise added to the TDC model decreases performance while more noise improves SM models. The additional noise may increase the diversity the low dispersion points, making them appear more like the TDC model.

\begin{table}[h!]
    \centering
    \begin{tabular}{@{}|l|l|l|l|@{}}
        \hline
        Dispersion & Noise & Acc & mIoU \\ 
        \hline
        No Points          & No Points     &  95.6   &  56.5    \\
        Low        & Low   & 95.9    &  59.6    \\
        Low        & High  &  96.9  &  61.0     \\
        High       & Low   &  \textbf{97.2}   &  \textbf{61.8}     \\
        High       & High  &  97.0   &  60.9    \\ 
        \hline
    \end{tabular}
    \caption{Comparison of accuracy and mIoU across crowdsourcing dispersion and noise levels. Low dispersion corresponds to a social media scraping approach, whereas high dispersion corresponds to a trained data collector approach.}
    \label{tab:points_ablation}
\end{table}

\section{Discussion and Conclusion}

Given the challenges with generating high quality labels for remote sensing data and the asset of community partners willing to participate in crowdsourcing, we sought to build an algorithm that utilized both modalities of observation to generate better flood segmentation masks for disaster resilience. We developed a multi-modal segmentation network trained on simple, easy to generate training data and synthetic crowdsourced data points. While we found that all types of crowdsourced data points improved a single-input segmentation model, the best results used data points dispersed across the perimeter of the event.

In practice, community members or government employees could provide data points along the perimeter of flood events with which we could train models to nowcast flood extents. Social media scraping, simple WhatsApp bots, and crowdsourcing-specific mobile applications could also be used to collect data and improve segmentation models. 

Future work should include a sensitivity analysis of the impact of crowdsourced points on the accuracy of Street to Cloud's predictions to determine how many points are necessary to outperform existing baselines for both crowdsourcing strategies. Additional studies of obtaining and parsing real crowdsourced data to determine the feasibility of both approaches is also required. Our success with a small volume of crowdsourced data suggests that coarse training labels could be generated using unsupervised or weakly supervised learning, which is another avenue to explore when determining how to deploy this algorithm in practice.

Street to Cloud is a prototype for a multi-modal segmentation network that uses crowdsourcing to mitigate the need for finely annotated training data. With further field testing across a variety of urban and rural domains and incorporating real crowdsourced data, we anticipate this model can be used widely to nowcast flood extents, monitor impacts, and inform disaster relief.

\bibliography{iclr2021_conference}
\bibliographystyle{iclr2021_conference}

\end{document}